%% file: bmvc_final.tex
\documentclass{bmvc2k}
\usepackage{graphicx}
\usepackage{amsmath}
\usepackage{amssymb}
\usepackage{amsthm}
\usepackage{algorithm}
\usepackage{algorithmicx}  
\usepackage{algpseudocode}
\usepackage{booktabs}
\usepackage{multirow}
\usepackage{colortbl}
\usepackage[switch]{lineno}
\usepackage{subcaption}
\usepackage{caption}
\usepackage[capitalize,noabbrev]{cleveref}
\usepackage{float}

\title{PADS: Plug-and-Play 3D Human Pose Analysis via Diffusion Generative Modeling}

\addauthor{Haorui Ji}{haorui.ji@anu.edu.au}{1}
\addauthor{Hongdong Li}{hongdong.li@anu.edu.au}{1}

\addinstitution{
 The Australian National University\\
 115 North Road \\
 Canberra, ACT, Australia
}

\runninghead{Haorui Ji, Hongdong Li}{PADS}

\def\etal{\emph{et al}\bmvaOneDot}

\begin{document}

\maketitle

\input{sec/0_abstract}
\input{sec/1_intro}
\input{sec/2_related_work}

\input{sec/3_method}
\input{sec/4_exp}
\input{sec/5_conclusion}

\bibliography{egbib}

\newpage
\appendix
\input{sec/X_suppl}

\end{document}

%% file: sec/0_abstract.tex
\begin{abstract}
Diffusion models have demonstrated impressive capabilities in modeling complex data distributions and are increasingly applied in various generative tasks. In this work, we propose \textbf{P}ose \textbf{A}nalysis by \textbf{D}iffusion \textbf{S}ynthesis (\textbf{PADS}), a unified generative modeling framework for 3D human pose analysis. PADS first learns a task-agnostic 3D pose prior via unconditional diffusion synthesis and then performs training-free adaptation to a wide range of pose analysis tasks, including 3D pose estimation, denoising, completion, etc., through a posterior sampling scheme. By formulating each task as an inverse problem with a known forward operator, PADS injects task-specific constraints during inference while keeping the pose prior fixed. This plug-and-play framework removes the need for task-specific supervision or retraining, offering flexibility and scalability across diverse conditions. Extensive experiments on different benchmarks showcase the superior performance against both learning-based and optimization-based baselines, demonstrating the effectiveness and generalization capability of our method.
\end{abstract}

%% file: sec/1_intro.tex
\section{Introduction}
\label{sec:intro}
3D human pose analysis (3D HPA) include a variety of tasks that aim to infer or reconstruct the 3D pose configuration of the human body under different conditions, such as estimating 3D poses from 2D observations, recovering missing joints, removing noise, or generating plausible samples. These tasks play an essential role in applications such as human-computer interaction, motion capture, healthcare, and virtual reality~\cite{jiang2022golfpose,weng2019photo,stenum2021applications}.

Most existing approaches focus on individual tasks and require task-specific architectures trained with paired supervision, such as 2D-3D pose correspondences. While effective, such methods suffer from two notable drawbacks: (1) limited generalization across task boundaries and require retraining or fine-tuning whenever task definitions or input modalities change; (2) method performance depends heavily on access to substantial volume of paired data.

\begin{figure}[t]
    \centering
    \includegraphics[width=0.7\linewidth]{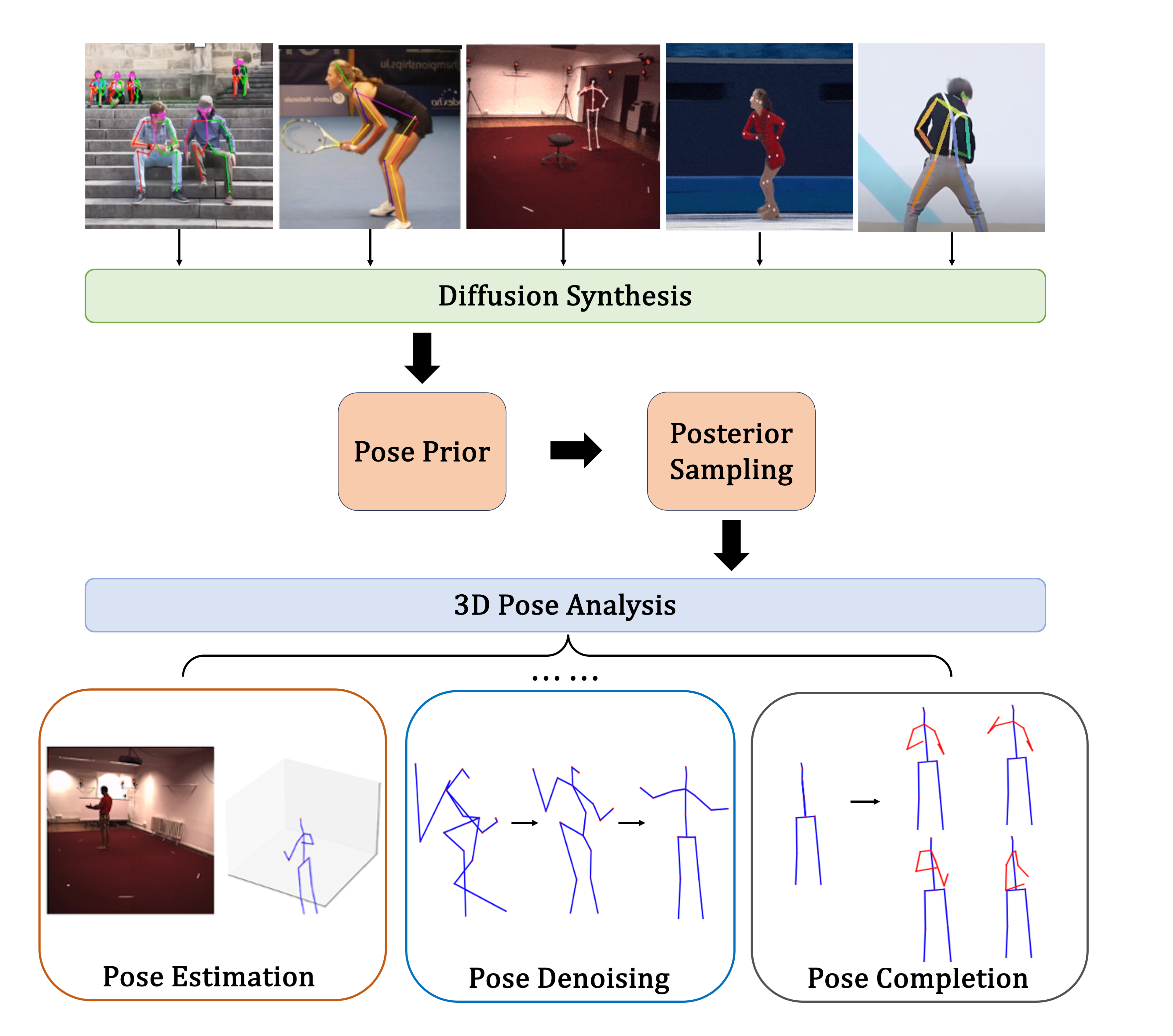}
    \caption{PADS utilize extensive human pose data to obtain an expressive and robust pose prior, which can be applied as a regularization term within posterior sampling scheme to discriminate infeasible solutions and address noise sensitivity. Its applications span across pose analysis tasks such as 3D pose estimation, denoising, completion.}
    \label{fig:pose prior example}
\end{figure}

To overcome these limitations, in this paper, we introduce \textbf{P}ose \textbf{A}nalysis by \textbf{D}iffusion \textbf{S}ynthesis (\textbf{PADS}), a general-purpose framework for 3D pose analysis that leverages the generative capabilities of diffusion models. The core idea of PADS consists of two aspects: (1) during training, it learns a task-agnostic 3D pose prior via unconditional diffusion synthesis, which captures the underlying kinematic constraints of human skeletons; and (2) at inference time, it addresses different downstream tasks by formulating them as inverse problems, where the learned prior is combined with a task-specific forward operator to guide the generation process through diffusion posterior sampling. This formulation enables a plug-and-play inference paradigm capable of handling diverse conditions such as noisy, partial, or 2D-observed poses, without the need of retraining or architectural modification. \cref{fig:pose prior example} illustrates the versatility of PADS across multiple use cases.

To summarize, our contributions are as follows:
\begin{itemize}
    \item We propose PADS, a unified generative framework that formulates diverse 3D human pose analysis tasks as inverse problems and solves them using pose prior and posterior sampling scheme.
    \item We demonstrate how PADS learns task-agnostic pose prior through diffusion synthesis, and how it enables training-free adaptation by injecting task-specific constraints at inference time via diffusion posterior sampling.
    \item We conduct extensive experiments on multiple benchmarks, showing competitive performance against both learning-based and optimization-based baselines.
\end{itemize}

%% file: sec/2_related_work.tex
\section{Related Work}
\label{sec:related work}
\subsection{3D Human Pose Analysis}
3D human pose analysis aims at reconstructing complete 3D human poses under different task conditions such as observed 2D pose, occluded 3D pose, pure noise, etc. Learning-based methods involve direct condition-to-pose learning through distinctive network designs. \cite{martinez2017simple,pavllo20193d} proposes a simple yet effective MLP-based neural network trained on 2D-3D poses pair. \cite{li2022mhformer,zhang2022mixste} utilize Transformer architecture to address sequential 3D pose analysis challenges. Other line of approaches take advantage of task-agnostic 3D pose priors such as parametric models\cite{loper2023smpl}, Gaussian mixtures\cite{bogo2016keep}, neural distance fields\cite{tiwari2022pose,he2024nrdf}, etc, and integrate them with task-specific constraints during the optimization process when applying to downstream tasks. Recently, diffusion models have been extensively applied in this research domain. Notable works include~\cite{jiang2023back,ji2023unsupervised}, which concentrate on solving pose estimation by leveraging manual designed loss to guide reverse diffusion process. \cite{ci2023gfpose} jointly learns the task-independent and task-aware pose priors in a supervised manner via hierarchical condition masking strategy. \cite{lu2023dposer} employ an optimization-based pipeline and use the learned prior as a regularization component to optimize for target poses. 

In this paper, PADS leverages the power of diffusion generative modeling while also enhancing the efficiency in inference stage by formulating downstream tasks as general inverse problems. Compared with similar frameworks~\cite{jiang2023back,ci2023gfpose,lu2023dposer}, we use skeleton instead of SMPL parameters as human pose representation, which offer more flexibility when modeling pose prior. In addition, our approach tackles different tasks in plug-and-play manner, enables easy adaptation to new scenarios without retraining.

\subsection{Diffusion Generative Modeling}
Diffusion models are a class of generative methods that model the underlying distribution over training samples. Their versatility in modeling complex data distributions has led to remarkable achievements across various research domains, including visual content generation~\cite{rombach2022high}, detection and segmentation~\cite{chen2023diffusiondet}. Recent applications have also expanded to 3D vision tasks, such as 3D reconstruction~\cite{poole2022dreamfusion,shan2023diffusion} and motion generation~\cite{tevet2022human,yuan2023physdiff}.

Recently, diffusion models have been extensively applied to inverse problems, particularly under the plug-and-play inference paradigm. In this context, inverse problems are formulated as sampling from conditional data distributions, with conditioning introduced via Bayes’ theorem. A variety of techniques~\cite{choi2021ilvr,kawar2022denoising,chung2022diffusion,song2022pseudoinverse} have been developed to incorporate observations into the sampling process, enabling the generation of samples from the target conditional distribution. However, the majority of existing work has focused on image processing tasks, such as inpainting, super-resolution, and deblurring, thereby limiting the generalization of these approaches to broader problem domains. In this work, we extend this framework to the domain of 3D human pose analysis, aiming to bridge this gap and demonstrate the broader applicability of diffusion-based inverse problem solvers.

%% file: sec/3_method.tex

\begin{figure}[t]
    \centering
    \includegraphics[width=\linewidth]{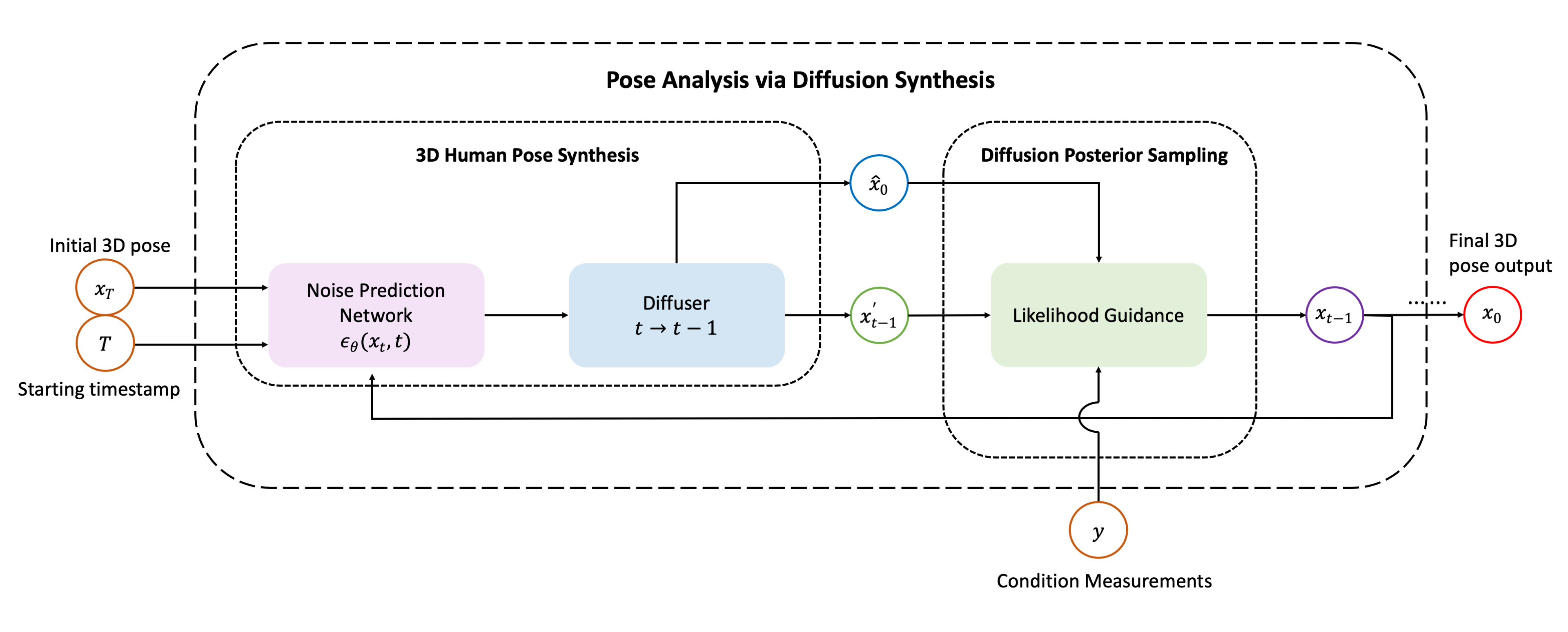}
    \caption{Overview of the PADS framework. During training, we train a noise prediction network for unconditional 3D human pose synthesis and model the underlying kinematic constraints. In the inference stage, we utilize the learned prior along with diffusion posterior sampling scheme to inject likelihood guidance into the sampling process.}
    \label{fig:pipeline}
\end{figure}

\section{Method}
\subsection{Problem Formulation}
\label{sec:problem formulation}
Various downstream tasks in 3D human pose analysis can be formulated as modeling conditional data distributions $p(\mathbf{x} | \mathbf{c})$ where $\mathbf{x} \in \mathbb{R}^{J \times 3}$ represents 3D human pose with $J$ skeleton joints, and $\mathbf{c}$ are different task conditions. In pose estimation, $\mathbf{c} \in \mathbb{R}^{J \times 2}$ corresponds to 2D poses, either from a single frame or a sequence, with the goal of reconstructing the 3D pose. In pose denoising and pose completion, $\mathbf{c} \in \mathbb{R}^{J \times 3}$ consists of noisy or partially occluded 3D poses, representing corrupted or incomplete skeletons. For pose generation, $\mathbf{c} \in \emptyset$ as the task involves unconstrained synthesis without external input.

\subsection{Pose Analysis as Inverse Problems}
At the core of PADS is the formulation of 3D pose analysis tasks as non-linear inverse problems, solved within a unified generative framework by sampling from target conditional distributions. A general inverse problem can be described as:
\begin{equation}
\label{inverse problem def}
    \mathbf{y} = f(\mathbf{x}) + \mathbf{n}, \quad \mathbf{x} \in \mathbb{R}^{n} \quad \mathbf{y},\mathbf{n} \in \mathbb{R}^{m},
\end{equation}
where $\mathbf{x}$ and $\mathbf{y}$ are the original signal and its corresponding measurement. $f(\cdot): \mathbb{R}^{n} \rightarrow \mathbb{R}^{m}$ is the forward operator and $\mathbf{n}$ denotes measurement noise. In the context of 3D pose analysis, $\mathbf{x}$ denotes the underlying 3D skeleton, $\mathbf{y}$ represents the task-specific condition, which is equivalent to $\mathbf{c}$ defined in \cref{sec:problem formulation}, and $f$ varies by task: for pose estimation, $f: \mathbb{R}^{3} \rightarrow \mathbb{R}^{2}$ represents perspective projection; for denoising, $f: \mathbb{R}^{3} \rightarrow \mathbb{R}^{3}$ applies noise perturbation; and for completion, $f: \mathbb{R}^{3} \rightarrow \mathbb{R}^{3}$ denotes a joint-wise masking operation.

\subsection{General Framework}
\label{pads}
Preliminaries for diffusion-based generative modeling are introduced in \cref{supp:preliminary} for reference. In the diffusion framework, sampling from a target distribution reduces to solving a reverse-time stochastic differential equation (SDE) using a learned score function. To enable conditional sampling, the original reverse SDE in \cref{eq:reverse SDE} can be modified as follows:
\begin{equation}
\label{eq:posterior SDE}
    \text{d} \mathbf{x} = [-\frac{\beta(t)}{2} \mathbf{x} - \beta(t) \nabla_{\mathbf{x}_t} \log p_t(\mathbf{x}_t | \mathbf{y})] \text{d} t + \sqrt{\beta(t)} \text{d} \mathbf{w}
\end{equation}
The posterior scores can be further decomposed via Bayes' theorem
\begin{equation}
\label{eq:bayes score}
    \nabla_{\mathbf{x}_t} \log p_t(\mathbf{x}_t | \mathbf{y})=\nabla_{\mathbf{x}_t} \log p_t(\mathbf{x}_t) + \nabla_{\mathbf{x}_t} \log p_t(\mathbf{y} | \mathbf{x}_t).
\end{equation}
where the first term corresponds to the prior score, estimated through unconditional diffusion modeling, and the second term reflects likelihood guidance to enforce data fidelity through the forward operator. Based on this decomposition, PADS adopts a two-stage pipeline: (1) during training, an unconditional diffusion model is learned to approximate the prior score function; and (2) during inference, this prior is combined with a task-specific inverse problem solver to incorporate likelihood guidance, ensuring consistency with observed conditions. \cref{fig:pipeline} illustrates a general framework that follows the strategy introduced above, and the following sections will dive into each component in more detail.

\subsection{Diffusion Pose Prior}
\label{pose diffusion synthesis}
The first step in our framework is to learn a robust, task-agnostic 3D human pose prior by training a diffusion-based synthesis network. To focus on plausible skeletal structures and eliminate spurious correlations with global orientation, we consider pelvis joint as root and synthesize the relative-to-pelvis 3D poses. By reconstructing input poses in a denoising autoencoding manner, we compel the diffusion model to learn effective kinematic constraints from genuine 3D poses.

To this end, we employ the denoising diffusion probabilistic model (DDPM)~\cite{ho2020denoising} for unconditional pose synthesis. Specifically, the forward process progressively corrupts the ground truth 3D pose $\mathbf{x}_0$ by adding Gaussian noise $\mathbf{\epsilon}_t$, producing a sequence $\mathbf{x}_1, \mathbf{x}_2, \dots \mathbf{x}_T$ that forms a Markov chain governed by the transition kernel $q(\mathbf{x}_{t} | \mathbf{x}_{t-1})$. Conversely, the reverse process then learns to denoise this sequence using a parameterized Gaussian kernel $p_\theta(\mathbf{x}_{t-1} | \mathbf{x}_t)$. 

Following~\cite{ho2020denoising}, we train a neural network $\mathbf{\epsilon}_\theta(\mathbf{x}_t, t)$ to predict the noise added at each timestep. The architecture is implemented using a Transformer encoder to capture long-range dependencies between joints, regardless of spatial structure. To align the learned reverse process with the true time reversal, we minimize the KL divergence between $p_\theta(\mathbf{x}_{t-1} | \mathbf{x}_t)$ and $q(\mathbf{x}_{t-1} | \mathbf{x}_t)$ at every timestamp, yielding the final loss function as:
\begin{equation}
    \begin{aligned}
        \mathcal{L} &= \mathbb{E}_{t, \mathbf{x}_0, \mathbf{\epsilon}_t}
        \left[ D_{KL}(q(\mathbf{x}_{t-1} | \mathbf{x}_t) || p_\theta(\mathbf{x}_{t-1} | \mathbf{x}_t)) \right] \\
        &= \mathbb{E}_{t, \mathbf{x}_0, \mathbf{\epsilon}_t}
        [\left\|\mathbf{\epsilon}_t-\mathbf{\epsilon}_\theta(\mathbf{x}_t, t)\right\|_2^2].
    \end{aligned}
\end{equation}
Note that~\cite{song2020score} has demonstrated that DDPM can also be interpreted as a discretized SDE governed by a learned score function. This unifies denoising and score-matching training paradigm, with the score function approximated as:
\begin{equation}
    \begin{aligned}
        \nabla_{\mathbf{x}_t} \log p_t(\mathbf{x}_t) = -\frac{1}{\sqrt{1-\bar{\alpha}_t}} \mathbf{\epsilon}_\theta(\mathbf{x}_t, t), \quad
        \bar{\alpha}_t = \prod_{i=1}^{t} (1 - \beta_t)
    \end{aligned}
\end{equation}

\subsection{Diffusion Posterior Sampling}
\label{dps}
Following the decomposition in \cref{eq:bayes score}, we incorporate likelihood guidance into the sampling process for conditional generation based on diffusion priors. To achieve this, we adopt the Diffusion Posterior Sampling (DPS)~\cite{chung2022diffusion} solver and generalize its formulation to 3D pose analysis. 

Specifically, DPS introduces an approximation quantified with the Jensen gap:
\begin{equation}
    \nabla_{\mathbf{x}_t} \log p(\mathbf{y} | \mathbf{x}_t) \simeq \nabla_{\mathbf{x}_t} \log p(\mathbf{y} | \hat{\mathbf{x}}_0) = -\frac{1}{\sigma^2} \nabla_{\mathbf{x}_t}\left\|\mathbf{y}-f(\hat{\mathbf{x}}_0)\right\|_2^2,
\end{equation}
$\hat{\mathbf{x}}_0 = \mathbb{E}[\mathbf{x}_0 | \mathbf{x}_t]$ denotes the closed-form posterior mean, $\sigma$ is the variance of measurement noise, and $f(\cdot)$ is the corresponding forward operator. Combining this with the learned pose prior from \cref{pose diffusion synthesis} yields,
\begin{equation}
\label{likelihood guidance}
    \nabla_{\mathbf{x}_t} \log p(\mathbf{x}_t | \mathbf{y}) \simeq -\frac{1}{\sqrt{1-\bar{\alpha}_t}} \mathbf{\epsilon}_{\theta}(\mathbf{x}_t, t)-\rho \nabla_{\mathbf{x}_t}\left\|\mathbf{y}-f(\hat{\mathbf{x}}_0)\right\|_2^2,
\end{equation}
where $\rho$ serves as a hyperparameter controlling the scale of the likelihood guidance.

In the inference stage, PADS applies the DPS method to 3D pose analysis, resulting in a training-free, plug-and-play paradigm capable of handling various downstream tasks without additional retraining. \cref{alg:pipeline} illustrates a pipeline for 3D human pose estimation task, and analogous pipelines for other tasks can be implemented by merely altering the measurement operator. Notably, pose estimation is inherently nonlinear due to the perspective projection. To align with the learned pose prior, we represent 3D poses relative to the pelvis, while the global trajectory of the root joint is provided separately. Inspired by~\cite{zheng2022truncated}, we propose a deterministic initialization strategy that replaces Gaussian noise with a camera-based inverse projection. Specifically, given the camera intrinsics $K$, 2D pose measurements $P_{2d}$, and the global root trajectory $T$, we construct an initial 3D skeleton by back-projecting each 2D joint ray and scaling it to match the root’s global displacement. This initialization produces a plausible 3D pose whose scale and coordinate frame are already consistent with the scene geometry.
\begin{equation}
    \begin{aligned}
        \widetilde{P}_{init} = \frac{K^{-1}P_{2d}}{\|K^{-1}P_{2d}\|_2} \|T\|_2, \quad
        P_{init} = \widetilde{P}_{init} - \widetilde{P}_{init}^{pelvis}.
    \end{aligned}
\end{equation}
Here, $\widetilde{P}_{init}$ corresponds to the back-projected and scaled 3D pose, while $P_{init}$ centers this pose at the pelvis to match the learned prior. By seeding diffusion with this geometry-aware initialization, the subsequent iterations focus on refining limb lengths and resolving projection nonlinearities, rather than discovering the global pose from scratch. As a result, sampling converges faster and with higher accuracy.

\begin{algorithm}[t]
\caption{PADS for 3D Human Pose Estimation}
\label{alg:pipeline}
\begin{algorithmic}[1]
\Require 
    \Statex 2D pose measurement $P_{2d}$; Camera intrinsic $K$; Pelvis root trajectory $T$;
    \Statex Pose prior $\mathbf{\epsilon}_{\theta}(\mathbf{x}_t, t)$; Number of diffusion sampling iterations $N$;
    \Statex Guidance scale $\left\{\rho_t\right\}_{t=1}^N$; Noise variance $\left\{\sigma_t\right\}_{t=1}^N$;

\State $\widetilde{P}_{init} \gets \frac{K^{-1}P_{2d}}{\|K^{-1}P_{2d}\|_2} \|T\|_2$ \Comment{Pose initialization}
\State $P_{init} \gets \widetilde{P}_{init} - \widetilde{P}_{init}^{pelvis}$

\State $\mathbf{x}_N \gets P_{init}$, $\mathbf{y} \gets P_{2d}$
\For{$t \gets N$ to $1$} \Comment{Iterative reverse sampling}
    \State $\hat{\mathbf{x}}_0 \gets \frac{1}{\sqrt{\bar{\alpha}_t}}(\mathbf{x}_t - \sqrt{1-\bar{\alpha}_t} \mathbf{\epsilon}_{\theta}(\mathbf{x}_t, t))$
    \State $\mathbf{z} \sim N(0, \mathbf{I})$
    \State $\mathbf{x}_{t-1}^{\prime} \gets \frac{\sqrt{\alpha_t}(1-\bar{\alpha}_{t-1})}{1-\bar{\alpha}_t} \mathbf{x}_t+\frac{\sqrt{\bar{\alpha}_{t-1}} \beta_t}{1-\bar{\alpha}_t} \hat{\mathbf{x}}_0+\sigma_t \mathbf{z}$
    \State $\mathbf{x}_{t-1} \gets \mathbf{x}_{t-1}^{\prime}-\rho_t \nabla_{\mathbf{x}_t}\left\|\mathbf{y}-proj(\hat{\mathbf{x}}_0)\right\|_2^2$
\EndFor
\State return $\mathbf{x}_0$
\end{algorithmic}
\end{algorithm}

%% file: sec/4_exp.tex
\section{Experiments}
\label{sec:exp}
This section presents the evaluation results of our proposed framework. We begin with an overview of experimental settings, including datasets and evaluation metrics. Subsequently, we showcase the quantitative and qualitative results for different pose analysis tasks, including estimation, generation, completion, and denoising. Due to page limits, we only report main results. Implementation details and ablation studies that investigate the impact of various design components.are reported in \cref{supp:implem details} and \cref{supp:quant results} for reference.

\subsection{Experiments Setup}
\noindent\textbf{Human3.6M}~\cite{ionescu2013human3} is the most widely used indoor benchmark for 3D human pose analysis. It captures a range of everyday activities from multiple camera views and provides high-precision 3D annotations via motion capture. Following common practice, we use subjects S1, S5, S6, S7, and S8 for training, and evaluate on subjects S9 and S11.

\noindent\textbf{MPI-INF-3DHP}~\cite{mehta2017monocular} is a more diverse dataset featuring more challenging scenes. It contains recordings of eight subjects performing various activities from 14 viewpoints. We evaluate on the valid frames specified by the official release.

\noindent\textbf{AMASS}~\cite{mahmood2019amass}. AMASS is a large dataset that provides detailed 3D body motion data, including SMPL~\cite{loper2023smpl} parameters, joint rotations, and body shapes. Following prior works~\cite{pavlakos2019expressive,lu2023dposer,tiwari2022pose}, we adopt the same train-test split and uniformly sample one-tenth of the original data for efficient training, yielding approximately two million pose samples.

\noindent\textbf{Evaluation metrics.} For pose estimation and reconstruction, we report the Mean Per Joint Position Error (MPJPE) and Procrustes-aligned MPJPE (PA-MPJPE) in millimeters, measuring the Euclidean distance between predicted and ground-truth joint positions. For MPI-INF-3DHP, we additionally report the Percentage of Correct Keypoints (PCK) at a 150mm threshold and the Area Under the Curve (AUC). To assess generation diversity, we report the Average Pairwise Distance (APD)~\cite{aliakbarian2020stochastic}, defined as the mean distance between all pairs of generated poses.

\begin{table*}[t]
\centering
\setlength{\abovecaptionskip}{0.2cm}
\resizebox{0.8\linewidth}{!}{%
    \begin{tabular}{ c  l  c  c  c  c }
    \toprule
        \multirow{2}{*}{ Mode } & \multirow{2}{*}{ Method } & \multicolumn{2}{c}{ GT } & \multicolumn{2}{c}{ DT } \\
        & & MPJPE $\downarrow$ & PA-MPJPE $\downarrow$ & MPJPE $\downarrow$ & PA-MPJPE $\downarrow$ \\
    \midrule
        \multirow{4}{*}{ Lrn. } 
        & Yu \etal \cite{yu2021towards} & 85.3 & 42.0 & 92.4 & 54.3 \\
        & PAUL \cite{wang2021paul} & 88.3 & - & 132.5 & - \\
        & MHR-Net \cite{zeng2022mhr} & 72.6 & - & - & - \\
        & ElePose \cite{wandt2022elepose} & \underline{64.0} & \underline{36.7} & - & - \\
    \midrule
        \multirow{5}{*}{ Opt. } 
        & SMPL \cite{bogo2016keep} & 82.3 & - & - & - \\
        & Song \etal \cite{song2020human} & - & 56.4 & - & - \\
        & ZeDO \cite{jiang2023back}~$(H=1)$ & - & - & 65.7 & 49.0 \\
        & ZeDO \cite{jiang2023back}~$(H=10)$ & - & - & \underline{57.1} & \underline{45.1} \\
        \rowcolor[gray]{0.85}
        & PADS & \textbf{41.5} & \textbf{33.1} & \textbf{54.8} & \textbf{44.9}\\
    \bottomrule
    \end{tabular}
}
\caption{Quantitative 3D HPE results on Human3.6M dataset. All results are reported in millimeters (mm). \textit{Lrn.} stands for learning-based methods and \textit{Opt.} means optimization-based methods. \textit{GT} and \textit{DT} respectively denote results using ground truth 2D inputs and detected 2D pose by off-the-shelf 2D detector. $H$ is the number of hypotheses. The best and second-best results are highlighted in bold and underline formats.}
\label{tab:h36m}
\vspace{-1em}
\end{table*}

\begin{table*}[htbp]
\centering
\setlength{\abovecaptionskip}{0.2cm}
\scalebox{0.9}{
    \begin{tabular}{ l  c  c  c}
    \toprule
    Method & PA-MPJPE $\downarrow$ & PCK $\uparrow$ & AUC $\uparrow$\\
    \midrule
        Chen \etal \cite{chen2019unsupervised} & - & 71.7 & 36.3\\
        Yu \etal \cite{yu2021towards} & - & \underline{86.2} & \underline{51.7} \\
        ElePose \cite{wandt2022elepose} & 54.0 & 86.0 & 50.1\\
        ZeDO \cite{jiang2023back} & 86.5 & 82.6 & 53.8 \\
        \rowcolor[gray]{0.85}
        PADS & \underline{63.7} & \textbf{86.6} & \textbf{57.6}\\
    \midrule
    \end{tabular}
}
\captionof{table}{Quantitative 3D HPE results on 3DHP dataset. Ground truth 2D inputs are used.}
\label{tab:3dhp}
\end{table*}

\begin{table*}[t]
\centering
\setlength{\abovecaptionskip}{0.01cm}
\resizebox{0.65\linewidth}{!}{%
    \begin{tabular}{ l  c  c  c}
    \toprule
    Method & GT-MPJPE $\downarrow$ & DT-MPJPE $\downarrow$ & \\
    \midrule
        DiffPose (CVPR2023)~\cite{gong2023diffpose} & \textbf{31.6} & 49.7 \\
        DiffPose (IROS2023)~\cite{choi2023diffupose} & 37.3 & 49.4 \\
        D3DP~\cite{shan2023diffusion} & \underline{34.24} & \textbf{44.3} \\
        DiffHPE~\cite{rommel2023diffhpe} & - & 51.2 \\
        PoseFormerV2~\cite{zhao2023poseformerv2} & 37.3 & \underline{47.5} \\
        MotionBERT~\cite{zhu2023motionbert} & 37.5 & - \\
        SPIN~\cite{kolotouros2019learning} & 41.1 & - \\
        HybrIK-X~\cite{li2025hybrik} & 47.0 & -\\
        \rowcolor[gray]{0.85}
        PADS & 41.5 & 54.8 \\
    \midrule
    \end{tabular}
}
\caption{Quantitative 3D HPE results on Human3.6M dataset with supervised methods. All methods employ single frame and single hypothesis setup.}
\label{tab:h36m additional}
\vspace{-0.5em}
\end{table*}

\subsection{Pose Estimation}
We first evaluate PADS on 3D human pose estimation (3D HPE), the core task in human pose analysis. PADS has three notable properties: (i) it requires no 2D–3D paired supervision during training; (ii) it enforces the learned pose prior as a constraint while iteratively optimizing the 3D pose; and (iii) it operates in a single-hypothesis setting, producing one 3D estimate per run. For fair comparison, results in \cref{tab:h36m} are benchmarked against methods with similar assumptions. PADS achieves 41.5 mm MPJPE and 33.1 mm PA-MPJPE, surpassing all state-of-the-art (SOTA) methods in this category by a clear margin. It also generalizes well under challenging conditions, maintaining robustness with both ground-truth and noisy 2D keypoints. On the 3DHP dataset, PADS attains 63.7 mm PA-MPJPE, 86.6 PCK, and 57.6 AUC (\cref{tab:3dhp}). While its PA-MPJPE lags behind the SOTA, PADS remains competitive in overall metrics against both learning- and optimization-based approaches, highlighting its effectiveness.

\begin{figure}[t]
\begin{center}
\includegraphics[width=\linewidth]{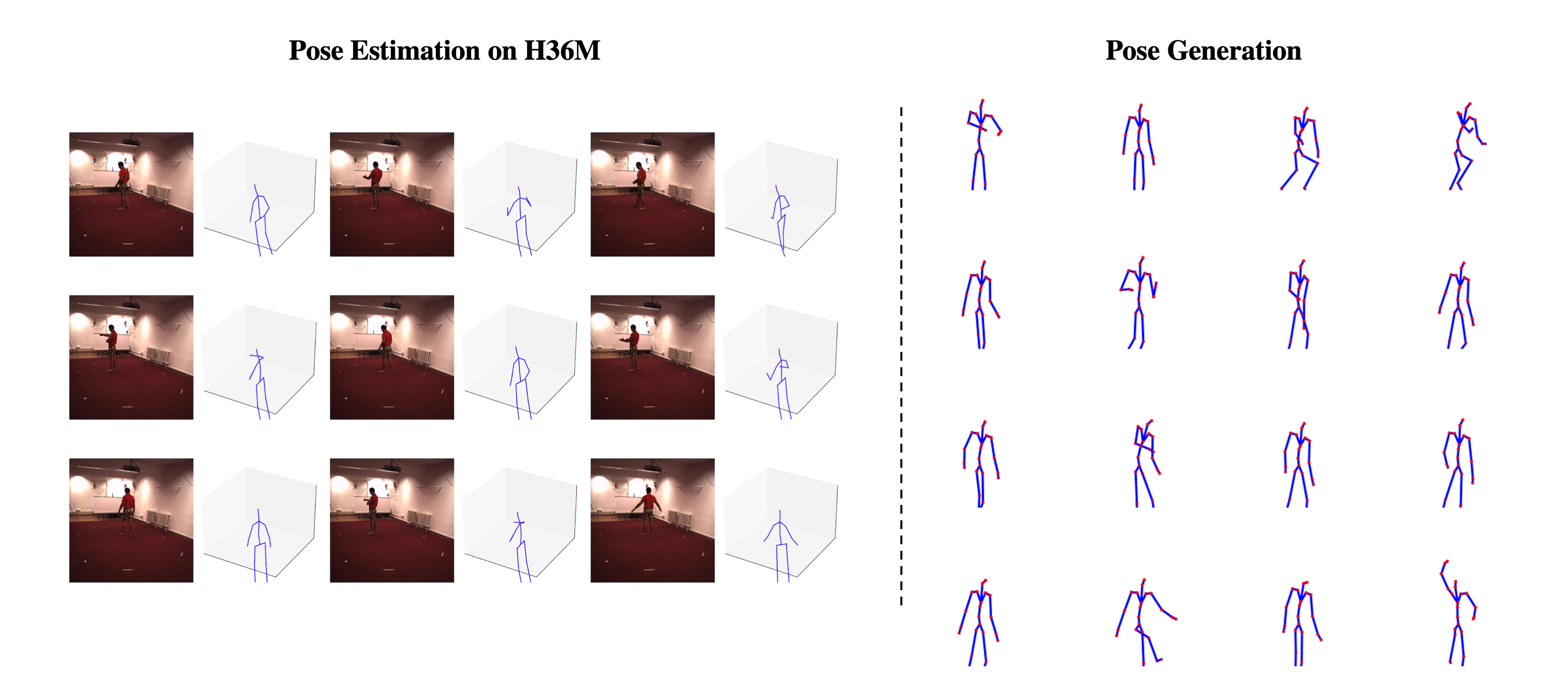}
\caption{Qualitative visualization of PADS's performance on different pose analysis tasks. Left is the pose estimation results on Human3.6M dataset, where each column constitutes an image depicting actors performing different actions and the predicted 3D pose. Right is the unconditional pose generation samples.}
\label{fig:visualization}
\end{center}
\end{figure}

To provide a broader perspective, we also compare PADS with fully supervised methods in \cref{tab:h36m additional}, including feedforward and diffusion-based baselines. Although PADS does not exceed these task-specific models, the performance gap is modest—despite requiring no paired supervision. This further underscores the efficiency of our framework. Qualitative results in \cref{fig:visualization} confirm the accuracy and coherence of the reconstructed 3D skeletons.

\subsection{Pose Generation}
For generation task, we train PADS on the AMASS dataset for unconditional human pose synthesis. We compare against representative baselines, including VPoser~\cite{pavlakos2019expressive}, Pose-NDF~\cite{tiwari2022pose}, and DPoser~\cite{lu2023dposer}, with quantitative and qualitative results summarized in \cref{tab:pose generation} and \cref{fig:visualization}. Following prior work, we randomly sample 500 poses and evaluate diversity using the APD metric. Since PADS represents poses as skeletons rather than meshes, direct comparison on shape realism is not feasible, highlighting the need for unified evaluation metrics applicable to both representations. Nevertheless, compared with AMASS ground truth and competing methods, PADS achieves competitive APD scores, demonstrating strong expressiveness and versatility both numerically and visually.

\subsection{Pose Completion} 
In real-world applications, 3D human poses obtained from wearable devices or vision-based algorithms often contain defects such as noise or missing joints. For instance, motion capture data may be corrupted by measurement errors, while avatars in metaverse applications may suffer from occlusion and yield partial observations. In the following sections, we demonstrate how PADS addresses such issues through pose completion and denoising. Both tasks can be formulated as linear inverse problems, enabling PADS to handle them in a unified manner.
For pose completion task, its forward measurement operator is formulated as joint-wise inpainting with a masking matrix, analogous to image inpainting:
\begin{equation}
    \mathbf{y} \sim \mathcal{N}\left(\mathbf{y} \mid \mathbf{M} \mathbf{x}, \sigma^2 \mathbf{I}\right)
\end{equation}
where $\mathbf{M} \in {0,1}^{n \times n}$ denotes the masking matrix. In our experiments, we mask groups of joints corresponding to semantic body parts (\textit{Leg}: hip, knee, foot; \textit{Arm}: shoulder, elbow, wrist; \textit{Spine}: hip, thorax, neck). Results in \cref{tab:pose completion} show that PADS reconstructs plausible poses from partially observed skeletons. Reconstruction of the upper body proves more challenging than the lower body, likely because arms are less constrained by the rest of the body and thus exhibit higher degrees of freedom. By contrast, spine reconstruction achieves superior performance, benefiting from stronger constraints imposed by surrounding joints.

\begin{table}[t]
\begin{minipage}[t]{0.47\linewidth}
\centering
\setlength{\abovecaptionskip}{0.2cm}
\scalebox{0.9}{
\begin{tabular}{ l  c }
    \toprule
     Sample source  & APD $\uparrow$ \\
    \midrule
        AMASS\cite{mahmood2019amass}  & 15.44  \\
    \midrule
        VPoser\cite{pavlakos2019expressive} & 10.75  \\
        PoseNDF\cite{tiwari2022pose} & \textbf{18.75}  \\
        DPoser\cite{lu2023dposer} & 14.28  \\
        \rowcolor[gray]{0.85}
        PADS & \underline{15.81} \\
    \bottomrule
    \end{tabular}
}
\captionof{table}{Quantitative comparison on pose generation task over 500 human poses.}
\label{tab:pose generation}
\end{minipage}
\hspace{0.5cm}
\begin{minipage}[t]{0.47\linewidth}
\centering
\setlength{\abovecaptionskip}{0.2cm}
\scalebox{0.8}{
\begin{tabular}{ c  c }
    \toprule
     Masked Body Parts  & MPJPE $\downarrow$ \\
    \midrule
        Right Leg & 17.1 \\
        Left Leg & 18.8 \\
        Right Arm & 32.6 \\
        Left Arm & 30.3 \\
        Two Legs & 63.3 \\
        Two Arms & 72.8\\
        Spine & 4.5\\
    \bottomrule
\end{tabular}
}
\captionof{table}{Pose completion results on Human3.6M dataset.}
\label{tab:pose completion}
\end{minipage}
\end{table}

\subsection{Pose Denoising}
For pose denoising, the forward operator follows the principle of diffusion process, perturbing inputs with predefined noise. Specifically, we inject different types of noise with varying intensity into the Human3.6M test set and evaluate PADS by measuring the MPJPE metric before and after denoising. As shown in \cref{tab:pose denoising}, PADS effectively mitigates all noise types across different intensities. With stronger perturbations, PADS requires more sampling steps, which is consistent with the design of the diffusion-based scheme.

\begin{table}[t]
\centering
\setlength{\abovecaptionskip}{0.2cm}
\scalebox{0.9}{
\begin{tabular}{ c  c  c  c }
    \toprule
     Noise Type & Intensity & Before$\downarrow$ & After$\downarrow$\\
    \midrule
        \multirow{3}{*}{ Gaussian }
        & $\mathcal{N}(0, 0.25)$ & 75.3 & 51.6 \\
        & $\mathcal{N}(0, 0.5)$ & 150.5 & 63.8 \\
        & $\mathcal{N}(0, 1.0)$ & 301.2 & 88.7 \\
    \midrule
        \multirow{3}{*}{ Uniform }        
        & $\mathcal{U}(-0.25, 0.25)$ & 45.2 & 43.1 \\
        & $\mathcal{U}(-0.5, 0.5)$ & 90.4 & 52.9 \\
        & $\mathcal{U}(-1.0, 1.0)$ & 180.9 & 69.6 \\
    \midrule     
        Poisson & $\mathcal{P}(0.25)$ & 152.3 & 90.1 \\
    \midrule     
        Salt and Pepper & $p = 0.25$ & 149.1 & 108.9 \\
    \bottomrule
\end{tabular}
}
\captionof{table}{Pose denoising results on Human3.6M dataset. We normalize the joint coordinates to [-1, 1] before injecting noise.}
\label{tab:pose denoising}
\end{table}

%% file: sec/5_conclusion.tex
\section{Conclusions}
This paper introduces PADS, a unified framework that formulates 3D human pose analysis as a class of inverse problems, enabling diverse downstream tasks to be solved within a single diffusion-based pipeline. At its core, PADS learns a task-agnostic 3D pose prior through unconditional diffusion, capturing the underlying kinematic structure. During inference, this prior is reused in a plug-and-play manner, where task-specific conditions are incorporated via posterior sampling, eliminating the need for task-specific retraining or paired supervision. Our results highlight the potential of diffusion-based frameworks to tackle a broad range of 3D pose tasks in a training-free setting. This study also points to several avenues for future work. One direction is extending the framework to richer human representations such as meshes or implicit functions. Another is developing a pose-specific solver tailored to skeletal structure and geometry, which may further improve both accuracy and efficiency.

%% file: sec/X_suppl.tex
\section{Preliminary: Diffusion Models}
\label{supp:preliminary}
Diffusion models are designed to implement generative modeling through a sequence of noising and denoising processes. \cite{song2020score} first defines a stochastic differential equation (SDE) that can transform complex data distributions to a known prior by progressively injecting noise. 
\begin{equation}
\label{eq:forward SDE}
    \text{d} \mathbf{x} = -\frac{\beta(t)}{2} \mathbf{x} \text{d} t + \sqrt{\beta(t)} d \mathbf{w}
\end{equation}
$\beta(t)$ is the schedule of the noising process, and $\mathbf{w}$ is the standard Wiener process. The original data distribution is defined at $t = 0$, i.e. $\mathbf{x}_0 \sim p_{data}$ , and a simple, tractable distribution (e.g. isotropic Gaussian) is achieved when $t = T$, i.e. $\mathbf{x}_T \sim \mathcal{N}(\mathbf{0}, \mathbf{I})$ when $T$ is sufficiently large.

Notably, original distributions can be restored by solving the corresponding reverse-time SDE from $t = T$ back to $t = 0$.
\begin{equation}
\label{eq:reverse SDE}
    \text{d} \mathbf{x} = \left[-\frac{\beta(t)}{2} \mathbf{x} - \beta(t) \nabla_{\mathbf{x}_t} \log p_t(\mathbf{x}_t)\right] \text{d} t+\sqrt{\beta(t)} \text{d} \mathbf{w}
\end{equation}
$\nabla_{\mathbf{x}_t} \log p_t(\mathbf{x}_t)$, also known as score functions, represents the gradient of the log probability densities with respect to data. The score function can be approximated with a neural network, which can be used to solve the reverse SDE in \cref{eq:reverse SDE} and generate novel samples once finishes training.

\section{Implementation Details}
\label{supp:implem details}
\begin{figure}[t]
    \centering
    \includegraphics[width=0.5\linewidth]{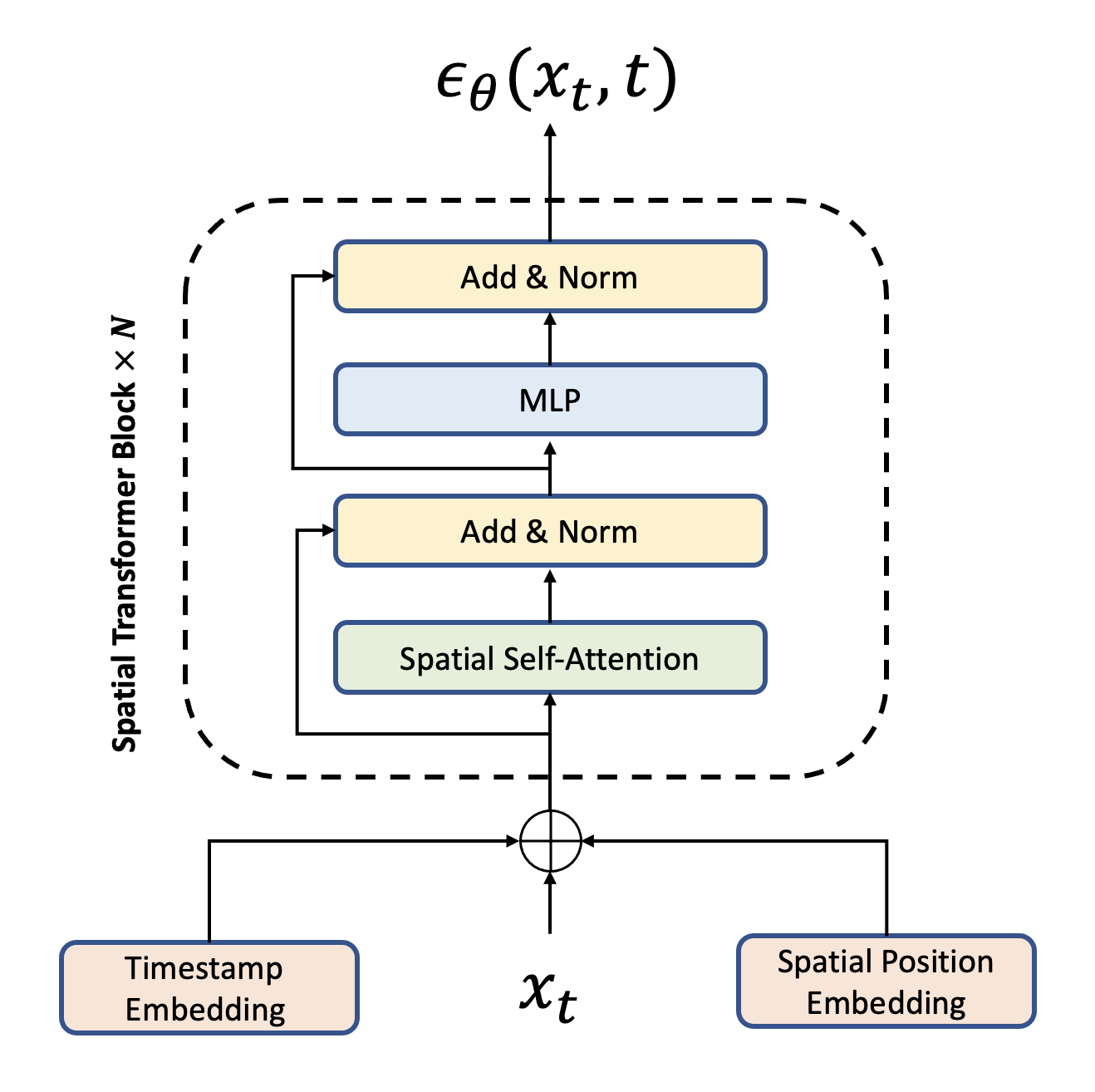}
    \caption{Transformer-based noise prediction network architecture.}
    \label{fig:network design}
\vspace{-0.5em}
\end{figure}

The architecture of Transformer-based noise prediction network $\mathbf{\epsilon}_\theta(\mathbf{x}_t, t)$ is illustrated in 
 \cref{fig:network design}. In the training stage, it is trained for 100 epochs with batch size of 1024, learning rate $10^{-4}$, and the AdamW optimizer. An exponential moving average strategy with ratio of 0.9999 is also adopted. For the diffusion process, we follow conventional DDPM training strategy, where the maximum iteration is set as $T=1000$, the timestamp $t$ is uniformly sampled from $[1, T]$, and the variances of added noises are configured to linearly increase from $\beta_1=10^{-4}$ to $\beta_T=0.02$. All 3D human poses are normalized to pelvis-related coordinates. Our experiments are performed on one NVIDIA 4070Ti GPU, and the training takes about 12 hours to complete.

During inference, as suggested in \cref{dps}, we truncate the complete diffusion chain to $T=450$ and iteratively sample from back to $t=0$ using the DDIM sampler~\cite{song2020denoising} in order to facilitate the sampling process. In every iteration, we fix the scale of the likelihood guidance to $\rho=0.003$.

\section{Additional Quantitative Results}
\label{supp:quant results}

\subsection{Ablation Studies on Pose Estimation}
\noindent\textbf{Effect of Diffusion Sampler.}
We begin by evaluating the performance of our method using different diffusion samplers. Specifically, we compare three sampling scheme: Score Matching~\cite{song2020score}, DDPM~\cite{ho2020denoising}, and DDIM~\cite{song2020denoising}. All models are trained under identical configurations and evaluated on the Human3.6M dataset. As shown in \cref{tab:diffbb}, DDIM achieves the best results in both MPJPE and PA-MPJPE. However, the other two samplers also yield competitive performance, comparable to many learning and optimization-based baselines. These results highlight the robustness and generality of our framework, which remains effective across different diffusion sampling strategies.

\noindent\textbf{Inverse Problem Solver Analysis.}
We further investigate the effect of different inverse problem solvers by replacing DPS with alternative state-of-the-art methods, including Manifold Constrained Gradients (MCG)~\cite{chung2022improving} and Pseudoinverse Guided Diffusion Models ($\Pi$GDM) \cite{song2022pseudoinverse}. For fairness, all solvers are applied to the same pretrained pose synthesis model. As shown in \cref{tab:inv solver}, the choice of solver significantly impacts the final performance, with DPS consistently outperforming the alternatives across all evaluation metrics. We attribute this to the specific characteristics of 3D human pose estimation and the non-linearity of the perspective projection operator. A deeper analysis of solver performance in relation to pose domain structure remains an interesting direction for future work.

\begin{table}[t]
\begin{minipage}[t]{0.45\linewidth}
\centering
\setlength{\abovecaptionskip}{0.2cm}
\scalebox{0.8}{
\begin{tabular}{ l  c  c }
    \toprule
     Diffusion Sampler  & MPJPE $\downarrow$ & PA-MPJPE $\downarrow$\\
    \midrule
      Score Matching~\cite{song2020score}   & 64.0  & 47.1 \\ 
      DDPM~\cite{ho2020denoising}   & 65.3  & 47.4 \\
      \rowcolor[gray]{0.85}
      DDIM~\cite{song2020denoising}   & \textbf{41.5} & \textbf{33.1}\\
    \bottomrule
    \end{tabular}
}
\captionof{table}{Quantitative 3D HPE results on Human3.6M dataset by using different diffusion sampler.}
\label{tab:diffbb}
\end{minipage}
\hspace{0.5cm}
\begin{minipage}[t]{0.45\linewidth}
\centering
\setlength{\abovecaptionskip}{0.2cm}
\scalebox{0.8}{
\begin{tabular}{ l  c  c }
    \toprule
     Inv. Problem Solver  & MPJPE $\downarrow$ & PA-MPJPE $\downarrow$\\
    \midrule
      MCG~\cite{chung2022improving}   & 80.0 & 56.1\\ 
      $\Pi$GDM~\cite{song2022pseudoinverse}  & 177.1  & 75.5\\
      \rowcolor[gray]{0.85}
      DPS~\cite{chung2022diffusion}   & \textbf{41.5} & \textbf{33.1}\\
    \bottomrule
    \end{tabular}
}
\captionof{table}{Quantitative 3D HPE results on Human3.6M dataset by using different diffusion-based inverse problem solver.}
\label{tab:inv solver}
\end{minipage}
\end{table}

\noindent\textbf{Pose Initialization Strategy Analysis.}
As described in \cref{dps}, we initialize the sampling process with the inverse projection of the input 2D pose, given the global trajectory of the root joint. We hypothesize that this strategy provides a more informative starting point than random noise, as it better aligns with the target 2D pose and reduces the number of diffusion steps required to generate an accurate 3D reconstruction. To validate this design choice, we conduct an ablation study on the Human3.6M dataset, examining how different initialization strategies and sampling iterations affect overall performance. We group configurations row-wise and employ a control variates approach to isolate the effect of each factor. As shown in \cref{tab:initialization}, the inverse projection strategy consistently outperforms random noise initialization, while also requiring fewer sampling steps to achieve comparable or better accuracy.
\begin{table}[t]
\setlength{\abovecaptionskip}{0.2cm}
\centering
\resizebox{0.6\linewidth}{!}{%
    \begin{tabular}{ c  c  c  c }
    \toprule
     Init. Strategy & Number Iter. & MPJPE $\downarrow$ & PA-MPJPE  $\downarrow$\\
    \midrule
        \multirow{4}{*}{ Random Noise }
        & 100 & 242.3 & 166.8 \\
        & 250 & 80.7 & 60.1 \\
        & 450 & 64.1 & 48.3 \\
        & 1000 & 79.4 & 54.6 \\
    \midrule
        \multirow{4}{*}{ Inverse Proj. }        
        & 100 & 54.6 & 39.8 \\
        & 250 & 42.6 & 34.5 \\
        & \cellcolor[gray]{0.85}450 & \cellcolor[gray]{0.85}\textbf{41.5} & \cellcolor[gray]{0.85}\textbf{33.1}\\
        & 1000 & 63.8 & 41.0 \\
    \bottomrule
    \end{tabular}
}
\caption{Quantitative 3D HPE results on Human3.6M dataset by using different pose initialization strategies and number of sampling iterations.}
\label{tab:initialization}
\end{table}

\subsection{Ablation Studies on Pose Denoising and Completion}
To verify the effectiveness of robustness of our design, we further ablate the influence of sampler and pose initialization strategy on denoising and completion tasks, as presented in \cref{supp tab:ablation}. According to the results, initializing from measurements outperforms from random noise meanwhile needing fewer iterations of sampling. In addition, DDIM sampler consistently outperforms DDPM.
\begin{table}[h]
\centering
\setlength{\abovecaptionskip}{0.2cm}
\resizebox{0.8\linewidth}{!}{%
    \begin{tabular}{ c  c  c  c  c  c}
    \toprule
         Task  &  Setup  &  Sampler & Init. Strategy & Number Iter. & MPJPE $\downarrow$ \\
    \midrule
    \multirow{4}{*}{ Denoising } 
        & GT + $\mathcal{N}(0, 0.25)$ & DDIM & Inverse Proj. & 150 & 51.6 \\
        & GT + $\mathcal{N}(0, 0.25)$ & DDIM & Inverse Proj. & 250 & 55.3 \\
        & GT + $\mathcal{N}(0, 0.25)$ & DDPM & Random Noise & 500 & 166.1 \\
        & GT + $\mathcal{N}(0, 0.25)$ & DDPM & Random Noise & 1000 & 158.8 \\
    \midrule
    \multirow{4}{*}{ Completion } 
        & Two Arms & DDIM & Inverse Proj. & 150 & 72.8 \\
        & Two Arms & DDIM & Inverse Proj. & 250 & 72.2 \\
        & Two Arms & DDPM & Random Noise & 500 & 93.9 \\
        & Two Arms & DDPM & Random Noise & 1000 & 92.8 \\
    \bottomrule
    \end{tabular}
}
\caption{Ablation studies on denoising and completion tasks. We ablate the influence of diffusion sampler, initialization strategy, and number of iterations.}
\label{supp tab:ablation}
\end{table}

%% file: bmvc_final.bbl
\begin{thebibliography}{48}
\providecommand{\natexlab}[1]{#1}
\providecommand{\url}[1]{\texttt{#1}}
\expandafter\ifx\csname urlstyle\endcsname\relax
  \providecommand{\doi}[1]{doi: #1}\else
  \providecommand{\doi}{doi: \begingroup \urlstyle{rm}\Url}\fi

\bibitem[Aliakbarian et~al.(2020)Aliakbarian, Saleh, Salzmann, Petersson, and Gould]{aliakbarian2020stochastic}
Sadegh Aliakbarian, Fatemeh~Sadat Saleh, Mathieu Salzmann, Lars Petersson, and Stephen Gould.
\newblock A stochastic conditioning scheme for diverse human motion prediction.
\newblock In \emph{Proceedings of the IEEE/CVF Conference on Computer Vision and Pattern Recognition}, pages 5223--5232, 2020.

\bibitem[Bogo et~al.(2016)Bogo, Kanazawa, Lassner, Gehler, Romero, and Black]{bogo2016keep}
Federica Bogo, Angjoo Kanazawa, Christoph Lassner, Peter Gehler, Javier Romero, and Michael~J Black.
\newblock Keep it smpl: Automatic estimation of 3d human pose and shape from a single image.
\newblock In \emph{Computer Vision--ECCV 2016: 14th European Conference, Amsterdam, The Netherlands, October 11-14, 2016, Proceedings, Part V 14}, pages 561--578. Springer, 2016.

\bibitem[Chen et~al.(2019)Chen, Tyagi, Agrawal, Drover, Mv, Stojanov, and Rehg]{chen2019unsupervised}
Ching-Hang Chen, Ambrish Tyagi, Amit Agrawal, Dylan Drover, Rohith Mv, Stefan Stojanov, and James~M Rehg.
\newblock Unsupervised 3d pose estimation with geometric self-supervision.
\newblock In \emph{Proceedings of the IEEE/CVF Conference on Computer Vision and Pattern Recognition}, pages 5714--5724, 2019.

\bibitem[Chen et~al.(2023)Chen, Sun, Song, and Luo]{chen2023diffusiondet}
Shoufa Chen, Peize Sun, Yibing Song, and Ping Luo.
\newblock Diffusiondet: Diffusion model for object detection.
\newblock In \emph{Proceedings of the IEEE/CVF International Conference on Computer Vision}, pages 19830--19843, 2023.

\bibitem[Choi et~al.(2023)Choi, Shim, and Kim]{choi2023diffupose}
Jeongjun Choi, Dongseok Shim, and H~Jin Kim.
\newblock Diffupose: Monocular 3d human pose estimation via denoising diffusion probabilistic model.
\newblock In \emph{2023 IEEE/RSJ International Conference on Intelligent Robots and Systems (IROS)}, pages 3773--3780. IEEE, 2023.

\bibitem[Choi et~al.(2021)Choi, Kim, Jeong, Gwon, and Yoon]{choi2021ilvr}
Jooyoung Choi, Sungwon Kim, Yonghyun Jeong, Youngjune Gwon, and Sungroh Yoon.
\newblock Ilvr: Conditioning method for denoising diffusion probabilistic models.
\newblock \emph{arXiv preprint arXiv:2108.02938}, 2021.

\bibitem[Chung et~al.(2022{\natexlab{a}})Chung, Kim, Mccann, Klasky, and Ye]{chung2022diffusion}
Hyungjin Chung, Jeongsol Kim, Michael~T Mccann, Marc~L Klasky, and Jong~Chul Ye.
\newblock Diffusion posterior sampling for general noisy inverse problems.
\newblock \emph{arXiv preprint arXiv:2209.14687}, 2022{\natexlab{a}}.

\bibitem[Chung et~al.(2022{\natexlab{b}})Chung, Sim, Ryu, and Ye]{chung2022improving}
Hyungjin Chung, Byeongsu Sim, Dohoon Ryu, and Jong~Chul Ye.
\newblock Improving diffusion models for inverse problems using manifold constraints.
\newblock \emph{Advances in Neural Information Processing Systems}, 35:\penalty0 25683--25696, 2022{\natexlab{b}}.

\bibitem[Ci et~al.(2023)Ci, Wu, Zhu, Ma, Dong, Zhong, and Wang]{ci2023gfpose}
Hai Ci, Mingdong Wu, Wentao Zhu, Xiaoxuan Ma, Hao Dong, Fangwei Zhong, and Yizhou Wang.
\newblock Gfpose: Learning 3d human pose prior with gradient fields.
\newblock In \emph{Proceedings of the IEEE/CVF Conference on Computer Vision and Pattern Recognition}, pages 4800--4810, 2023.

\bibitem[Gong et~al.(2023)Gong, Foo, Fan, Ke, Rahmani, and Liu]{gong2023diffpose}
Jia Gong, Lin~Geng Foo, Zhipeng Fan, Qiuhong Ke, Hossein Rahmani, and Jun Liu.
\newblock Diffpose: Toward more reliable 3d pose estimation.
\newblock In \emph{Proceedings of the IEEE/CVF Conference on Computer Vision and Pattern Recognition}, pages 13041--13051, 2023.

\bibitem[He et~al.(2024)He, Tiwari, Birdal, Lenssen, and Pons-Moll]{he2024nrdf}
Yannan He, Garvita Tiwari, Tolga Birdal, Jan~Eric Lenssen, and Gerard Pons-Moll.
\newblock Nrdf: Neural riemannian distance fields for learning articulated pose priors.
\newblock In \emph{Proceedings of the IEEE/CVF Conference on Computer Vision and Pattern Recognition}, pages 1661--1671, 2024.

\bibitem[Ho et~al.(2020)Ho, Jain, and Abbeel]{ho2020denoising}
Jonathan Ho, Ajay Jain, and Pieter Abbeel.
\newblock Denoising diffusion probabilistic models.
\newblock \emph{Advances in neural information processing systems}, 33:\penalty0 6840--6851, 2020.

\bibitem[Ionescu et~al.(2013)Ionescu, Papava, Olaru, and Sminchisescu]{ionescu2013human3}
Catalin Ionescu, Dragos Papava, Vlad Olaru, and Cristian Sminchisescu.
\newblock Human3. 6m: Large scale datasets and predictive methods for 3d human sensing in natural environments.
\newblock \emph{IEEE transactions on pattern analysis and machine intelligence}, 36\penalty0 (7):\penalty0 1325--1339, 2013.

\bibitem[Ji et~al.(2023)Ji, Deng, Dai, and Li]{ji2023unsupervised}
Haorui Ji, Hui Deng, Yuchao Dai, and Hongdong Li.
\newblock Unsupervised 3d pose estimation with non-rigid structure-from-motion modeling.
\newblock \emph{arXiv preprint arXiv:2308.10705}, 2023.

\bibitem[Jiang et~al.(2022)Jiang, Ji, Menaker, and Hwang]{jiang2022golfpose}
Zhongyu Jiang, Haorui Ji, Samuel Menaker, and Jenq-Neng Hwang.
\newblock Golfpose: Golf swing analyses with a monocular camera based human pose estimation.
\newblock In \emph{2022 IEEE International Conference on Multimedia and Expo Workshops (ICMEW)}, pages 1--6. IEEE, 2022.

\bibitem[Jiang et~al.(2023)Jiang, Zhou, Li, Chai, Yang, and Hwang]{jiang2023back}
Zhongyu Jiang, Zhuoran Zhou, Lei Li, Wenhao Chai, Cheng-Yen Yang, and Jenq-Neng Hwang.
\newblock Back to optimization: Diffusion-based zero-shot 3d human pose estimation.
\newblock \emph{arXiv preprint arXiv:2307.03833}, 2023.

\bibitem[Kawar et~al.(2022)Kawar, Elad, Ermon, and Song]{kawar2022denoising}
Bahjat Kawar, Michael Elad, Stefano Ermon, and Jiaming Song.
\newblock Denoising diffusion restoration models.
\newblock \emph{Advances in Neural Information Processing Systems}, 35:\penalty0 23593--23606, 2022.

\bibitem[Kolotouros et~al.(2019)Kolotouros, Pavlakos, Black, and Daniilidis]{kolotouros2019learning}
Nikos Kolotouros, Georgios Pavlakos, Michael~J Black, and Kostas Daniilidis.
\newblock Learning to reconstruct 3d human pose and shape via model-fitting in the loop.
\newblock In \emph{Proceedings of the IEEE/CVF international conference on computer vision}, pages 2252--2261, 2019.

\bibitem[Li et~al.(2025)Li, Bian, Xu, Chen, Yang, and Lu]{li2025hybrik}
Jiefeng Li, Siyuan Bian, Chao Xu, Zhicun Chen, Lixin Yang, and Cewu Lu.
\newblock Hybrik-x: Hybrid analytical-neural inverse kinematics for whole-body mesh recovery.
\newblock \emph{IEEE Transactions on Pattern Analysis and Machine Intelligence}, 2025.

\bibitem[Li et~al.(2022)Li, Liu, Tang, Wang, and Van~Gool]{li2022mhformer}
Wenhao Li, Hong Liu, Hao Tang, Pichao Wang, and Luc Van~Gool.
\newblock Mhformer: Multi-hypothesis transformer for 3d human pose estimation.
\newblock In \emph{Proceedings of the IEEE/CVF Conference on Computer Vision and Pattern Recognition}, pages 13147--13156, 2022.

\bibitem[Loper et~al.(2023)Loper, Mahmood, Romero, Pons-Moll, and Black]{loper2023smpl}
Matthew Loper, Naureen Mahmood, Javier Romero, Gerard Pons-Moll, and Michael~J Black.
\newblock Smpl: A skinned multi-person linear model.
\newblock In \emph{Seminal Graphics Papers: Pushing the Boundaries, Volume 2}, pages 851--866. 2023.

\bibitem[Lu et~al.(2023)Lu, Lin, Dou, Zhang, Deng, and Wang]{lu2023dposer}
Junzhe Lu, Jing Lin, Hongkun Dou, Yulun Zhang, Yue Deng, and Haoqian Wang.
\newblock Dposer: Diffusion model as robust 3d human pose prior.
\newblock \emph{arXiv preprint arXiv:2312.05541}, 2023.

\bibitem[Mahmood et~al.(2019)Mahmood, Ghorbani, Troje, Pons-Moll, and Black]{mahmood2019amass}
Naureen Mahmood, Nima Ghorbani, Nikolaus~F Troje, Gerard Pons-Moll, and Michael~J Black.
\newblock Amass: Archive of motion capture as surface shapes.
\newblock In \emph{Proceedings of the IEEE/CVF international conference on computer vision}, pages 5442--5451, 2019.

\bibitem[Martinez et~al.(2017)Martinez, Hossain, Romero, and Little]{martinez2017simple}
Julieta Martinez, Rayat Hossain, Javier Romero, and James~J Little.
\newblock A simple yet effective baseline for 3d human pose estimation.
\newblock In \emph{Proceedings of the IEEE international conference on computer vision}, pages 2640--2649, 2017.

\bibitem[Mehta et~al.(2017)Mehta, Rhodin, Casas, Fua, Sotnychenko, Xu, and Theobalt]{mehta2017monocular}
Dushyant Mehta, Helge Rhodin, Dan Casas, Pascal Fua, Oleksandr Sotnychenko, Weipeng Xu, and Christian Theobalt.
\newblock Monocular 3d human pose estimation in the wild using improved cnn supervision.
\newblock In \emph{2017 international conference on 3D vision (3DV)}, pages 506--516. IEEE, 2017.

\bibitem[Pavlakos et~al.(2019)Pavlakos, Choutas, Ghorbani, Bolkart, Osman, Tzionas, and Black]{pavlakos2019expressive}
Georgios Pavlakos, Vasileios Choutas, Nima Ghorbani, Timo Bolkart, Ahmed~AA Osman, Dimitrios Tzionas, and Michael~J Black.
\newblock Expressive body capture: 3d hands, face, and body from a single image.
\newblock In \emph{Proceedings of the IEEE/CVF conference on computer vision and pattern recognition}, pages 10975--10985, 2019.

\bibitem[Pavllo et~al.(2019)Pavllo, Feichtenhofer, Grangier, and Auli]{pavllo20193d}
Dario Pavllo, Christoph Feichtenhofer, David Grangier, and Michael Auli.
\newblock 3d human pose estimation in video with temporal convolutions and semi-supervised training.
\newblock In \emph{Proceedings of the IEEE/CVF conference on computer vision and pattern recognition}, pages 7753--7762, 2019.

\bibitem[Poole et~al.(2022)Poole, Jain, Barron, and Mildenhall]{poole2022dreamfusion}
Ben Poole, Ajay Jain, Jonathan~T Barron, and Ben Mildenhall.
\newblock Dreamfusion: Text-to-3d using 2d diffusion.
\newblock \emph{arXiv preprint arXiv:2209.14988}, 2022.

\bibitem[Rombach et~al.(2022)Rombach, Blattmann, Lorenz, Esser, and Ommer]{rombach2022high}
Robin Rombach, Andreas Blattmann, Dominik Lorenz, Patrick Esser, and Bj{\"o}rn Ommer.
\newblock High-resolution image synthesis with latent diffusion models.
\newblock In \emph{Proceedings of the IEEE/CVF conference on computer vision and pattern recognition}, pages 10684--10695, 2022.

\bibitem[Rommel et~al.(2023)Rommel, Valle, Chen, Khalfaoui, Marlet, Cord, and P{\'e}rez]{rommel2023diffhpe}
C{\'e}dric Rommel, Eduardo Valle, Micka{\"e}l Chen, Souhaiel Khalfaoui, Renaud Marlet, Matthieu Cord, and Patrick P{\'e}rez.
\newblock Diffhpe: Robust, coherent 3d human pose lifting with diffusion.
\newblock In \emph{Proceedings of the IEEE/CVF International Conference on Computer Vision}, pages 3220--3229, 2023.

\bibitem[Shan et~al.(2023)Shan, Liu, Zhang, Wang, Han, Wang, Ma, and Gao]{shan2023diffusion}
Wenkang Shan, Zhenhua Liu, Xinfeng Zhang, Zhao Wang, Kai Han, Shanshe Wang, Siwei Ma, and Wen Gao.
\newblock Diffusion-based 3d human pose estimation with multi-hypothesis aggregation.
\newblock \emph{arXiv preprint arXiv:2303.11579}, 2023.

\bibitem[Song et~al.(2020{\natexlab{a}})Song, Meng, and Ermon]{song2020denoising}
Jiaming Song, Chenlin Meng, and Stefano Ermon.
\newblock Denoising diffusion implicit models.
\newblock \emph{arXiv preprint arXiv:2010.02502}, 2020{\natexlab{a}}.

\bibitem[Song et~al.(2022)Song, Vahdat, Mardani, and Kautz]{song2022pseudoinverse}
Jiaming Song, Arash Vahdat, Morteza Mardani, and Jan Kautz.
\newblock Pseudoinverse-guided diffusion models for inverse problems.
\newblock In \emph{International Conference on Learning Representations}, 2022.

\bibitem[Song et~al.(2020{\natexlab{b}})Song, Chen, and Hilliges]{song2020human}
Jie Song, Xu~Chen, and Otmar Hilliges.
\newblock Human body model fitting by learned gradient descent.
\newblock In \emph{European Conference on Computer Vision}, pages 744--760. Springer, 2020{\natexlab{b}}.

\bibitem[Song et~al.(2020{\natexlab{c}})Song, Sohl-Dickstein, Kingma, Kumar, Ermon, and Poole]{song2020score}
Yang Song, Jascha Sohl-Dickstein, Diederik~P Kingma, Abhishek Kumar, Stefano Ermon, and Ben Poole.
\newblock Score-based generative modeling through stochastic differential equations.
\newblock \emph{arXiv preprint arXiv:2011.13456}, 2020{\natexlab{c}}.

\bibitem[Stenum et~al.(2021)Stenum, Cherry-Allen, Pyles, Reetzke, Vignos, and Roemmich]{stenum2021applications}
Jan Stenum, Kendra~M Cherry-Allen, Connor~O Pyles, Rachel~D Reetzke, Michael~F Vignos, and Ryan~T Roemmich.
\newblock Applications of pose estimation in human health and performance across the lifespan.
\newblock \emph{Sensors}, 21\penalty0 (21):\penalty0 7315, 2021.

\bibitem[Tevet et~al.(2022)Tevet, Raab, Gordon, Shafir, Cohen-Or, and Bermano]{tevet2022human}
Guy Tevet, Sigal Raab, Brian Gordon, Yonatan Shafir, Daniel Cohen-Or, and Amit~H Bermano.
\newblock Human motion diffusion model.
\newblock \emph{arXiv preprint arXiv:2209.14916}, 2022.

\bibitem[Tiwari et~al.(2022)Tiwari, Anti{\'c}, Lenssen, Sarafianos, Tung, and Pons-Moll]{tiwari2022pose}
Garvita Tiwari, Dimitrije Anti{\'c}, Jan~Eric Lenssen, Nikolaos Sarafianos, Tony Tung, and Gerard Pons-Moll.
\newblock Pose-ndf: Modeling human pose manifolds with neural distance fields.
\newblock In \emph{European Conference on Computer Vision}, pages 572--589. Springer, 2022.

\bibitem[Wandt et~al.(2022)Wandt, Little, and Rhodin]{wandt2022elepose}
Bastian Wandt, James~J Little, and Helge Rhodin.
\newblock Elepose: Unsupervised 3d human pose estimation by predicting camera elevation and learning normalizing flows on 2d poses.
\newblock In \emph{Proceedings of the IEEE/CVF Conference on Computer Vision and Pattern Recognition}, pages 6635--6645, 2022.

\bibitem[Wang and Lucey(2021)]{wang2021paul}
Chaoyang Wang and Simon Lucey.
\newblock Paul: Procrustean autoencoder for unsupervised lifting.
\newblock In \emph{Proceedings of the IEEE/CVF Conference on Computer Vision and Pattern Recognition}, pages 434--443, 2021.

\bibitem[Weng et~al.(2019)Weng, Curless, and Kemelmacher-Shlizerman]{weng2019photo}
Chung-Yi Weng, Brian Curless, and Ira Kemelmacher-Shlizerman.
\newblock Photo wake-up: 3d character animation from a single photo.
\newblock In \emph{Proceedings of the IEEE/CVF conference on computer vision and pattern recognition}, pages 5908--5917, 2019.

\bibitem[Yu et~al.(2021)Yu, Ni, Xu, Wang, Zhao, and Zhang]{yu2021towards}
Zhenbo Yu, Bingbing Ni, Jingwei Xu, Junjie Wang, Chenglong Zhao, and Wenjun Zhang.
\newblock Towards alleviating the modeling ambiguity of unsupervised monocular 3d human pose estimation.
\newblock In \emph{Proceedings of the IEEE/CVF International Conference on Computer Vision}, pages 8651--8660, 2021.

\bibitem[Yuan et~al.(2023)Yuan, Song, Iqbal, Vahdat, and Kautz]{yuan2023physdiff}
Ye~Yuan, Jiaming Song, Umar Iqbal, Arash Vahdat, and Jan Kautz.
\newblock Physdiff: Physics-guided human motion diffusion model.
\newblock In \emph{Proceedings of the IEEE/CVF International Conference on Computer Vision}, pages 16010--16021, 2023.

\bibitem[Zeng et~al.(2022)Zeng, Yu, Miao, and Yang]{zeng2022mhr}
Haitian Zeng, Xin Yu, Jiaxu Miao, and Yi~Yang.
\newblock Mhr-net: Multiple-hypothesis reconstruction of non-rigid shapes from 2d views.
\newblock In \emph{Computer Vision--ECCV 2022: 17th European Conference, Tel Aviv, Israel, October 23--27, 2022, Proceedings, Part II}, pages 1--17. Springer, 2022.

\bibitem[Zhang et~al.(2022)Zhang, Tu, Yang, Chen, and Yuan]{zhang2022mixste}
Jinlu Zhang, Zhigang Tu, Jianyu Yang, Yujin Chen, and Junsong Yuan.
\newblock Mixste: Seq2seq mixed spatio-temporal encoder for 3d human pose estimation in video.
\newblock In \emph{Proceedings of the IEEE/CVF conference on computer vision and pattern recognition}, pages 13232--13242, 2022.

\bibitem[Zhao et~al.(2023)Zhao, Zheng, Liu, Wang, and Chen]{zhao2023poseformerv2}
Qitao Zhao, Ce~Zheng, Mengyuan Liu, Pichao Wang, and Chen Chen.
\newblock Poseformerv2: Exploring frequency domain for efficient and robust 3d human pose estimation.
\newblock In \emph{Proceedings of the IEEE/CVF conference on computer vision and pattern recognition}, pages 8877--8886, 2023.

\bibitem[Zheng et~al.(2022)Zheng, He, Chen, and Zhou]{zheng2022truncated}
Huangjie Zheng, Pengcheng He, Weizhu Chen, and Mingyuan Zhou.
\newblock Truncated diffusion probabilistic models and diffusion-based adversarial auto-encoders.
\newblock \emph{arXiv preprint arXiv:2202.09671}, 2022.

\bibitem[Zhu et~al.(2023)Zhu, Ma, Liu, Liu, Wu, and Wang]{zhu2023motionbert}
Wentao Zhu, Xiaoxuan Ma, Zhaoyang Liu, Libin Liu, Wayne Wu, and Yizhou Wang.
\newblock Motionbert: A unified perspective on learning human motion representations.
\newblock In \emph{Proceedings of the IEEE/CVF International Conference on Computer Vision}, pages 15085--15099, 2023.

\end{thebibliography}
